\documentclass{article}
\usepackage{graphicx}
\usepackage{hyperref}

\begin{document}

\title{Evolving Evolutionary Algorithms with Patterns}

\author{Mihai Oltean\\
Department of Computer Science\\
Faculty of Mathematics and Computer Science\\
Babe\c s-Bolyai University, Kog\u alniceanu 1\\
Cluj-Napoca, 3400, Romania.\\
\\
mihai.oltean@gmail.com
}

\date{}
\maketitle

\begin{abstract}

A new model for evolving Evolutionary Algorithms (EAs) is proposed in this paper. The model is based on the Multi Expression Programming (MEP) technique. Each MEP chromosome encodes an evolutionary pattern which is repeatedly used for generating the individuals of a new generation. The evolved pattern is embedded into a standard evolutionary scheme which is used for solving a particular problem. Several evolutionary algorithms for function optimization are evolved by using the considered model. The evolved evolutionary algorithms are compared with a human-designed Genetic Algorithm. Numerical experiments show that the evolved evolutionary algorithms can compete with standard approaches for several well-known benchmarking problems.

\end{abstract}


\section{Introduction}

Solving problems defines the human history. Unfortunately, each problem requires its own method for solving it. Modern techniques and algorithms tried to overcome this drawback. For instance, Evolutionary Computation \cite{goldberg1,holland1} tries to shift the difficulty from algorithms to representation: the algorithm is intend to be a general one, whereas the solution representation and associated genetic operators should be problem dependent.

The questions related to the efficiency of a particular evolutionary scheme have led the birth of No Free Lunch theorems for Search and Optimization \cite{wolpert1,wolpert2}. The answer was negative: we cannot have the best Evolutionary Algorithm (EA) which could perfectly solve all the optimization problems. All we can do is designing optimal or near-optimal algorithms for some particular problems without expecting any guaranty related to their generalization ability on new and unseen test problems.

Many evolutionary algorithms have been proposed for dealing with optimization problems. However, the lack of mathematical proofs of convergence does exist in most of the cases. Only some experimental evidence has been employed for supporting the proposed theories.

Instead of involving sophisticated mathematical theories we propose a simple way of deriving good evolutionary algorithms for a particular problem or set of problems: by evolving them. Our purpose is to evolve an evolutionary algorithm capable of solving a particular optimization problem. Thus, we will work with EAs at two levels: the first (macro) level consists in a steady-state EA \cite{syswerda1} which uses a fixed population size, a fixed mutation probability, a fixed crossover 
probability etc. The second (micro) level consists in the solutions encoded in a chromosome of the first level EA.

The macro level evolves a population of evolutionary algorithms in the micro level. The evolutionary algorithms in the micro level evolve solutions for a particular optimization problem (such as function optimization, TSP, etc.) Thus the output of the micro level EAs are solutions for a particular problem being solved, whereas the output of macro level EA is the best evolutionary algorithm in the micro level.

\begin{figure*}[htbp]
\centerline{\includegraphics[width=\textwidth]{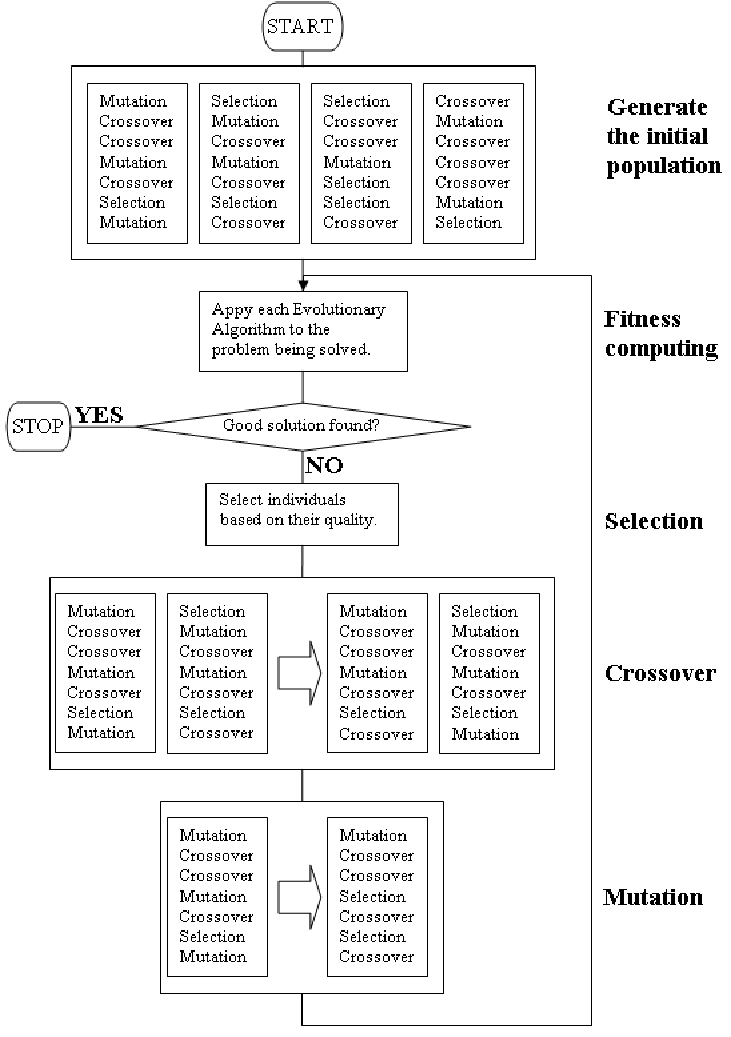}} 
\caption{A schematic view of the process of evolving evolutionary algorithms. Macro level EA evolves evolutionary algorithms within micro level. The output of the macro-level EA is the best EA within micro-level.}
\label{fig_mep_eas}
\end{figure*}

For the first (macro) level EA we use a standard evolutionary model, similar to 
Multi Expression Programming (MEP) \cite{oltean_complex,oltean_ecal,oltean_parity}, which is very suitable for 
evolving computer programs that may be easily translated into an imperative 
language (like \textbf{\textit{C}} or \textbf{\textit{Pascal}}). The basic idea of evolving evolutionary algorithms is depicted in Figure \ref{fig_mep_eas}.

Instead of evolving an entire EA we evolve only the heart of the algorithm, which is the sequence of instructions that generates a new offspring by taking information from the current population. The size of the search space is considerable smaller in this case. Numerical experiments performed with the proposed model reveal a short training time for obtaining a good evolutionary algorithm capable of competing a standard genetic algorithm for some particular test problems.

Note that the rules employed by the evolved EAs are not preprogrammed. These rules are automatically discovered by the evolution.

This research was motivated by the need of answering several important 
questions concerning evolutionary algorithms:\\

\textit{What is the optimal structure of an evolutionary algorithm?}\\

\textit{Which are the genetic operators that have to be used in conjunction with an EA (for a 
given problem)?}\\

\textit{Which is the optimal (or near-optimal) sequence of genetic operations (selections, 
crossovers and mutations) to be performed during a generation of an 
EA for a particular problem?}\\

For instance, in a standard genetic algorithm the sequence is the following: selection, recombination and mutation. But, how do we know that this scheme is the best for a particular problem (or problem instance)? We had better let the evolution to find the answer for us. Note that evolving only the probabilities for applying specific genetic operators is not enough. Also, the structure of an EA and the order in which these genetic operators are applied is also important.

The source code for evolving Evolutionary Algorithms is available at \url{https://github.com/mihaioltean/evolve-algorithms}.

The paper is organized as follows. An overview of the related work in the field of evolving evolutionary algorithms is made in section \ref{review}. The new model proposed for evolving EAs is presented in section \ref{proposed}. 
Several numerical experiments are performed in section \ref{experiments}. These experiments proved that the model proposed in this paper is able to evolve good evolutionary algorithm capable of competing with human-written algorithms. Further research directions are suggested in section \ref{future}.

\section{Related Work}\label{review}

Several attempts for evolving Evolutionary Algorithms using similar techniques were made in the past. A 
non-generational EA was evolved \cite{oltean_ecal} by using the Multi Expression 
Programming (MEP) technique \cite{oltean_ecal,oltean_parity}. A generational EA was evolved \cite{oltean_lgp,oltean_mit} by using the Linear Genetic Programming (LGP) technique \cite{brameier1,nordin1}. Numerical experiments have shown \cite{oltean_ecal,oltean_lgp} that the evolved EAs perform similarly and sometimes even better than the standard evolutionary approaches with which they are compared.

There are also several approaches that evolve genetic operators for solving 
difficult problems \cite{angeline1,angeline2,edmonds1,stephens1,teller1}. In his paper on Meta-Genetic Programming, Edmonds \cite{edmonds1} used two populations: a standard GP population and a co-evolved population of operators that act on the main population. Note that all these approaches use a fixed evolutionary algorithm which is not changed during the search.

Attempts for evolving complex heuristics for particular problems were made in the recent past. In \cite{oltean_tsp} the authors evolve an heuristic for the Traveling Salesman Problem. The obtained heuristic is a mathematical expression that takes as input some information about the already constructed path and outputs the next node of the path. It was shown \cite{oltean_tsp} that the evolved heuristic performs better than other well-known heuristics (Nearest Neighbor Heuristic, Minimum Spanning Tree Heuristic \cite{cormen1,garey1}) for the considered test problems.

A recent paper of Spector and Robinson \cite{spector1} describes a language called Push which supports a new, self-adaptive form of evolutionary computation called autoconstructive evolution. An experiment showing good results was reported for symbolic regression problems.

\subsection{Multi Expression Programming based Approach}
\label{mep_based}

Multi Expression Programming (MEP) \cite{oltean_complex,oltean_ecal,oltean_parity} \footnote{MEP source code is available at \url{https://mepx.org} or \url{https://github.com/mepx}.} uses a special representation that is similar to the way in which C and Pascal compilers translate mathematical expressions into machine code \cite{aho1}. The MEP genes are (represented by) substrings of variable length. The number of genes in a chromosome is constant and it represents the chromosome length. Each gene encodes a terminal (an element in the terminal set $T$) or a function symbol (an element in the function set $F$). A gene encoding a function includes pointers towards the function arguments. Function parameters always have indices of lower values than the position of that function itself in the chromosome.

According to the proposed representation scheme, the first symbol in a chromosome must be a terminal symbol. In this way only syntactically correct programs are obtained.
\\
\\
{\bf Example}

We use a representation where the numbers on the left stand for gene labels (or memory addresses). Labels do not belong to the chromosome, they are provided only for explanatory purposes. An example of a chromosome is given below (assuming that $T$ = \{$a$, $b$, $c$, $d$\} and $F$ = \{+, -, *, /\}):\\

1: $a$

2: $b$

3: + 1, 2

4: $c$

5: $d$

6: + 4, 5

7: * 2, 6\\

The MEP chromosomes are read top-down, starting with the first position. A terminal symbol specifies a simple expression. A function symbol specifies a complex expression (formed by linking the operands specified by the argument positions with the current function symbol). 

The expressions encoded in this chromosome are: \\

$E_1 = a$;

$E_2 = b$;

$E_3 = a + b$;

$E_4 = c$;

$E_5 = d$;

$E_6 = c + d$;

$E_7 = b * (c + d)$.\\

We have to choose one of these expressions ($E_1,\dots,E_7$) to represent the chromosome. There is neither theoretical nor practical evidence that one of them is better than the others. Thus, we choose to encode multiple solutions in a single chromosome. Each MEP chromosome encodes a number of expressions equal to the chromosome length (the number of genes). The expression associated to each chromosome position is obtained by reading the chromosome bottom-up from the current position, by following the links provided by the function pointers. The fitness of each expression encoded in a MEP chromosome is computed in a conventional manner (the fitness depends on the problem being solved). The best expression encoded in an MEP chromosome is chosen to represent the chromosome (the fitness of an MEP individual equals the fitness of the best expression encoded in that chromosome).

Genetic operators used in conjunction with MEP are crossover and mutation (see \cite{oltean_complex} for more information).

\subsubsection{Evolving full EAs with MEP}
\label{mep_full_eas}

In order to use MEP for evolving EAs the set of terminal symbols and a set of function symbols have been redefined \cite{oltean_ecal}. We have to keep in mind that the value stored in a terminal symbol is independent of other symbols in the chromosome and a function symbol changes the solution stored in another gene. 

We have decided to use 4 types of genetic operators in our evolved EA. The evolution will decide which operators are good and which are bad for our problem. The employed operators along with their meaning are given below:\\

\begin{itemize}
\item {\it Initialization} - randomly initializes a solution,
\item {\it Selection} - selects the best solution among several already existing solutions
\item {\it Crossover} - recombines two already existing solutions,
\item {\it Mutation} - varies an already existing solution.\\
\end{itemize}

These operators will act as symbols that may appear into an MEP chromosome. The only operator that generates a solution independent of the already existing solutions is the {\it Initialization} operator. This operator will constitute the terminal set. The other operators will be considered function symbols. Thus, we have $T$ = \{{\it Initialization}\}, $F$ = \{{\it Selection}, {\it Crossover}, {\it Mutation}\}.

An MEP chromosome $C$, storing an evolutionary algorithm, can look like:\\
\\
{\tt
1: {\it Initialization}\hspace{1.2cm}\{Randomly generates a solution\}\\
2: {\it Initialization}\hspace{1.2cm}\{Randomly generates another solution\}\\
3: {\it Mutation} 1\hspace{1,45cm}\{Mutates the solution stored on position 1\}\\
4: {\it Selection} 1, 3\hspace{1cm}\{Selects the best solution from those\}\\
\hspace*{4,5cm}\{stored on positions 1 and 3\}\\
5: {\it Crossover} 2, 4\hspace{0,75cm}\{Recombines the solutions on positions 2 and 4\}\\ 
6: {\it Mutation} 4\hspace{1,6cm}\{Mutates the solution stored on position 4\}\\
7: {\it Mutation} 5\hspace{1,6cm}\{Mutates the solution stored on position 5\}\\
8: {\it Crossover} 2, 6\hspace{0,79cm}\{Recombines the solutions on positions 2 and 6\}\\
}

This MEP chromosome encodes multiple evolutionary algorithms (in fact 8 EAs as many). They are given in Table \ref{mep_solutions}. Each EA is obtained by reading the chromosome bottom-up, starting with the current gene and following the links provided by the function pointers. The best evolutionary algorithm encoded into a chromosome will represent that chromosome (will provide the fitness of that chromosome). 

The complexity of the EAs encoded into a MEP chromosome varies from very simple (EAs made up of a single instruction) to very complex (sometimes using all genes of the MEP chromosome). This is very useful because we do not know, in advance, the complexity of the EA required to solve a problem. The required algorithm could be very simple (and, in this case, the simplest individuals encoded by MEP are very useful) or it could be very complex (and, in this case, the most complex EAs are taken into account).

Thus we deal with EAs at two different levels: a micro level representing the evolutionary algorithm encoded in an MEP chromosome and a macro level GA, which evolves MEP individuals. The number of genetic operators (initializations, crossovers, mutations, selections) is not fixed and it may vary between 1 and the MEP chromosome length. These values are automatically discovered by the evolution. The macro level GA execution is bound by the known rules for GAs (\cite{goldberg1}). This process is depicted in Figure \ref{fig_mep_eas}.

\begin{table*}
\begin{center}\caption{Evolutionary Algorithms encoded in the MEP chromosome $C$}
\label{mep_solutions}
\begin{tabular}{|l|l|l|l|}
\hline
$EA_1$ & $EA_2$ & $EA_3$ & $EA_4$\\
\hline
$i_1$={\it Initialization} & $i_1$={\it Initialization} & $i_1$={\it Initialization} & $i_1$ = {\it Initialization}\\
 & & $i_2$={\it Mutation}($i_1$) & $i_2$={\it Mutation}($i_1$)\\
 & & & $i_3$={\it Selection}($i_1$,$i_2$)\\
\hline
$EA_5$ & $EA_6$ & $EA_7$ & $EA_8$\\
\hline
$i_1$={\it Initialization} & $i_1$={\it Initialization} & $i_1$={\it Initialization} & $i_1$={\it Initialization}\\
$i_2$={\it Initialization} & $i_2$={\it Mutation}($i_1$) & $i_2$={\it Initialization} & $i_2$={\it Initialization}\\
$i_3$={\it Mutation}($i_1$) & $i_3$={\it Selection}($i_1, i_2$) & $i_3$={\it Mutation}($i_1$) & $i_3$={\it Mutation}($i_1$)\\
$i_4$={\it Selection}($i_1, i_3$) & $i_4$={\it Mutation}($i_3$) & $i_4$={\it Selection}($i_1,i_3$) & $i_4$={\it Selection}($i_1,i_3$)\\
$i_5$={\it Crossover}($i_1,i_4$) & & $i_5$={\it Crossover}($i_2,i_4$) & $i_5$={\it Mutation}($i_4$)\\
 & & $i_6$={\it Mutation}($i_5$) & $i_6$={\it Crossover}($i_2,i_5$)\\
\hline
\end{tabular}
\end{center}
\end{table*}

\paragraph{Remarks}

\begin{itemize}
\item[$(i)$]In our model, the {\it Crossover} operator always generates a single offspring from two parents. The crossover operators generating two offspring may also be designed to fit our evolutionary model.
\item[$(ii)$]The {\it Selection} operator acts as a binary tournament selection. The better out of two individuals is always accepted as the selection result.
\item[$(iii)$]	The {\it Initialization, Crossover} and {\it Mutation} operators are problem dependent. 
\end{itemize}

\subsubsection {Fitness assignment}
\label{mep_fitness}

We have to compute the quality of each EA encoded in the chromosome in order to establish the fitness of an MEP individual. For this purpose each EA encoded in an MEP chromosome is run on the particular problem being solved.

Roughly speaking the fitness of an MEP individual is equal to the fitness of the best solution generated by one of the evolutionary algorithms encoded in that MEP chromosome. But, since the EAs encoded in an MEP chromosome use pseudo-random numbers it is likely that successive runs of the same EA generate completely different solutions. This stability problem is handled in the following manner: each EA encoded in an MEP chromosome was executed (run) more times and the fitness of an MEP chromosome is the average of the fitness of the best EA encoded in that chromosome over all runs. In all the experiments performed in \cite{oltean_ecal} each EA encoded into an MEP chromosome was run 200 times.

\subsection{Linear Genetic Programming based Approach}

\textit{Linear Genetic Programming} (LGP) \cite{banzhaf1,brameier1,nordin1} uses a specific linear representation of computer programs. Programs of an imperative language (like 
\textbf{\textit{C}}) are evolved instead of the tree-based GP expressions of a functional programming language 
(like \textbf{\textit{LISP}}). 

An LGP individual is represented by a variable-length sequence of simple 
\textbf{\textit{C}} language instructions. Instructions operate on one or 
two indexed variables (registers) $r$ or on constants $c$ from predefined sets. 
The result is assigned to a destination register, e.g. $r_{i}=r_{j}$ * $c$. 

An example of an LGP program is the following:\\

\textsf{\textbf{void}}\textsf{ 
LGP{\_}Program(}\textsf{\textbf{double}} $v[8]$)

\textsf{{\{}}

\textsf{\ldots }

\hspace{0.5cm}\textsf{$v[0] = v[5] + 73;$}

\hspace{0.5cm}\textsf{$v[7] = v[4] - 59;$}

\hspace{0.5cm}\textsf{$v[4] = v[2] *v[1];$}

\hspace{0.5cm}\textsf{$v[2] = v[5] + v[4];$}

\hspace{0.5cm}\textsf{$v[6] = v[1] * 25;$}

\hspace{0.5cm}\textsf{$v[6] = v[4] - 4;$}

\hspace{0.5cm}\textsf{$v[1] = sin(v[6]);$}

\hspace{0.5cm}\textsf{$v[3] = v[5] * v[5];$}

\hspace{0.5cm}\textsf{$v[7] = v[6] * 2;$}

\hspace{0.5cm}\textsf{$v[5] = [7] + 115;$}

\hspace{0.5cm}\textsf{$v[1] = sin(v[7]);$}

\textsf{{\}}}\\

\subsubsection{LGP for Evolving EAs}

The structure of an LGP chromosome has been adapted for evolving EAs \cite{oltean_lgp}. Instead of working with registers, our LGP program will modify an array of individuals (the population). We denote by \textit{Pop} the array of 
individuals (the population) which will be modified by an LGP program. 

The set of function symbols will consist in genetic operators that may 
appear into an evolutionary algorithm. There are usually 3 types of genetic 
operators that may appear into an EA (see section \ref{mep_based}): \textit{Crossover, Mutation and Selection}. The \textit{Initialization} operator (see section \ref{mep_based}) is used for initialization of the population and it is not evolved like all other operators.

The LGP statements are considered to be genetic operations executed 
during an EA generation. Since the purpose was to evolve a generational EA 
a wrapper loop has been added around the genetic operations that are executed 
during an EA generation. Even more, each EA starts with a random 
population of individuals. Thus, the LGP program must contain some 
instructions that initialize the initial population. 

An LGP chromosome encoding an EA is given below:\\

\textsf{\textbf{void}}\textsf{ 
LGP{\_}Program(}\textsf{\textbf{Chromosome}}
\textsf{ $Pop[8]$) // a population with of 8 individuals}

\textsf{{\{}}

\hspace{0.5cm}\textsf{Randomly{\_}initialize{\_}the{\_}population();}

\hspace{0.5cm}\textsf{// repeat for a number of generations}

\hspace{0.5cm}\textsf{\textbf{for}}\textsf{ (}\textsf{\textbf{int}}\textsf{ $k$ = 0; $k$ $<$ 
$MaxGenerations$; $k$++){\{} }

\hspace{1cm}\textsf{$Pop[0]$ = $Mutation$($Pop[5]$);}

\hspace{1cm}\textsf{$Pop[7]$ = $Selection$($Pop[3]$, $Pop[6]$);}

\hspace{1cm}\textsf{$Pop[4]$ = $Mutation$($Pop[2]$);}

\hspace{1cm}\textsf{$Pop[2]$ = $Crossover$($Pop[0]$, $Pop[2]$);}

\hspace{1cm}\textsf{$Pop[6]$ = $Mutation$($Pop[1]$);}

\hspace{1cm}\textsf{$Pop[2]$ = $Selection$($Pop[4]$, $Pop[3]$);}

\hspace{1cm}\textsf{$Pop[1]$ = $Mutation$($Pop[6]$);}

\hspace{1cm}\textsf{$Pop[3]$ = $Crossover$($Pop[5]$, $Pop[1]$);}

\hspace{1cm}\textsf{{\}}}

\textsf{{\}}}\\

\textbf{\textit{Remarks}}

\begin{itemize}

\item[{\it (i)}]{The initialization function and the \textbf{for} 
cycle will not be affected by the genetic operators. These parts are kept 
unchanged during the search process.}

\item[{\it (ii)}]{Each LGP chromosome encodes a single EA. This is different from the MEP approach where each chromosome encodes multiple EAs.}

\end{itemize}

The quality of an EA encoded in an LGP chromosome is computed as in the case of the MEP approach (see section \ref{mep_fitness}). The EA encoded in an LGP chromosome is run on the particular problem being solved and the quality of the EA is equal to the quality of the best solution evolved by that EA.

\section{Proposed Model}\label{proposed}

The model proposed for evolving evolutionary algorithms is described in this section.

\subsection{Motivation}

This section tries to answer an important question:\\

\textit{"Why evolving EAs with patterns?"}\\

In the previously described models (MEP-based and LGP-based), the search space of the evolutionary algorithms was huge. The time needed to train a human-competitive evolutionary algorithm could take between several hours and several days.

Instead of evolving an entire EA we will try to evolve a small piece of code that will repeatedly be used in order to obtain new individuals. Most of the known evolutionary schemes use this form of evolution. For instance, in a standard GA, the following piece of code is successively applied until the new population is filled:\\

$p_1$ = $Selection$ (); // randomly choose two individuals from the current population and 

\hspace{3cm}//the best of them will be the result of the selection

$p_2$ = $Selection$ (); // choose another two randomly individuals and return the best of them in $p_2$

$c$ = $Crossover$($p_1$, $p_2$);

$c$ = $Mutation$($c$);

$Fitness$($c$);           // compute the fitness of the individual $c$

$Copy$ $c$ in the next generation;\\

The patern employed by a simple (1+1) Evolution Strategy \cite{beyer1} is:\\

$b$ = $Mutation$($a$);

$Fitness$($b$)         // compute the fitness of the individual $b$

\textbf{if} $b$ is better than $a$

\textbf{then} $Replace$ $a$ with $b$;

\textbf{endif}\\

The patern employed by a Steady-State Evolutionary Algorithm \cite{syswerda1} is:\\

$p_1$ = $Selection$ (); // randomly choose two individuals from the current population and 

\hspace{3cm}//the best of them will be the result of the selection

$p_2$ = $Selection$ (); // choose another two randomly individuals and return the best of them in $p_2$

$c$ = $Crossover$($p_1$, $p_2$);

$c$ = $Mutation$($c$);

$Fitness$($c$);        // compute the fitness of the individual $c$

\textbf{if} $c$ is better than the worst individual in the population
	
\textbf{then} $Replace$ the worst individual in the population with $c$;

\textbf{endif}\\

The main advantage of this approach is its reduced complexity: the size of the pattern is considerably smaller than the size of the entire EA as evolved in \cite{oltean_ecal,oltean_lgp}.

The patterns in an evolutionary algorithm can be assimilated with the Automatically Defined Functions (ADFs) in Genetic Programming \cite{koza2}. 

\subsection{Representation}

As shown in the previous section, the use of patterns greatly reduces the size of the search space.

Our pattern will be represented as an MEP computer program whose instructions will be executed during the EA evolution. We have chosen Multi Expression Programming for representing the patterns because MEP provides an easy way to store a sequence of instructions. Note that, in our approach, MEP will store only one solution (pattern) per chromosome because, due to the generational algorithm where the pattern will be embedded, is difficult to handle multiple evolutionary patterns in the same chromosome. We will still use the MEP notation but the there will be only one solution per chromosome in all cases and experiments below.

Other GP methods (such as standard Genetic Programming \cite{koza1} and Gene Expression Programming \cite{ferreira1} may also be used for storing patterns. Linear Genetic Programming \cite{brameier1} is less suitable for this model because we are not very interested in storing (by using some destination variables) the results of the crossover, selection and mutation operators. However, in our implementation we use some destination variables, but these variables are fixed and they are not subject to evolution (as in the LGP model).

As previously described in section \ref{mep_based}, we have decided to use 4 types of genetic operation within an EA:\\

\begin{itemize}
\item {\it Initialization} - randomly initializes a solution,
\item {\it Selection} - selects the best solution among several already existing solutions,
\item {\it Crossover} - recombines two already existing solutions,
\item {\it Mutation} - varies an already existing solution.\\
\end{itemize}

Since our purpose is to evolve a small piece of code that will be used in order to generate a new population based on the old population we will not use the \textit{Initialization} operator. This operator is used once at the moment of population initialization. The other three operators \textit{Selection, Crossover} and \textit{Mutation} will be used within the evolved pattern.

The operators that appear in the evolved EA and their new meanings are shown below:\\

\begin{itemize}
\item {\it Selection} - selects a solution from the old population. This operation is implemented as a binary tournament selection: two individuals are randomly chosen and the best of them is the result of selection.
\item {\it Crossover (a, b)} - recombines solutions $a$ and $b$.
\item {\it Mutation (a)} - varies solution $a$.\\
\end{itemize}

Function set is $F$ = \{$Crossover$, $Mutation$\} and the terminal set is $T$ = \{$Selection$\}. As a restriction imposed by the MEP representation, the first instruction in a chromosome must be a $Selection$.\\

Let us recall the MEP representation as given in section \ref{mep_based}. A MEP chromosome can have the form (we took only the first 5 genes of the chromosome given in section \ref{mep_based}):\\

1: $a$

2: $b$

3: + 1, 2

4: * 4, 5

5: $c$\\

We will replace the terminal set \{$a$, $b$, ...\} by the new terminal symbol \{$Selection$\} which is specific to our purpose. Also the function set \{+, *, -, ...\} will be replaced by \{$Crossover$, $Mutation$\}.\\

\textbf{Example 1}\\

An example of a MEP chromosome encoding a pattern is given below:\\

1: $Selection$

2: $Selection$

3: $Mutation$ 1

4: $Mutation$ 2

5: $Crossover$ 2, 4\\

This MEP chromosome should be interpreted as follows:\\

\begin{itemize}
\item{An individual (let us denote it by $a$) is selected from the current population}
\item{Another individual (let us denote it by $b$) is selected from the current population}
\item{Individual $a$ is mutated. The result of the mutation is a new individual denoted by $c$.}
\item{Individual $b$ is mutated. A new individual (denoted by $d$) is obtained.}
\item{Individuals $b$ and $d$ are recombined using a problem-dependent crossover. A new individual $e$ is obtained.}\\
\end{itemize}

\textbf{Example 2}\\

The pattern employed by a standard GA rewritten as an MEP chromosome is given below:\\

1: $Selection$

2: $Selection$

3: $Crossover$ 1, 2

4: $Mutation$ 3\\

The obtained individual is added to the new population.

\subsection{What brings new the proposed approach}

There are several important differences between the MEP-based technique \cite{oltean_ecal} presented in section \ref{mep_full_eas} and the one proposed in this paper. First of all, the complexity of the evolved algorithms is quite different. In the first approach \cite{oltean_ecal} we have evolved an entire EA, which is a simple sequence of genetic instructions (no generations, no loops for filling a population). In the current approach we keep intact most of the GA parts: we have generations and we have a loop which fills the new population.

Another important modifications have been performed to the genetic operators employed by the evolved evolutionary algorithms. As a general modification all genetic operators manipulate only individuals from the current population. This is again different from the method proposed in \cite{oltean_ecal} whose operators manipulate individuals taken from the set of all individuals created since the beginning of the search process.

Other particular modifications are described below:\\

\begin{itemize}

\item{$Initialization$. In the MEP-based approach proposed in \cite{oltean_ecal} we could use initialization operator anywhere in the evolved algorithm. Here, in the current approach, this operator has been removed from the set of possible operators that could appear into an evolved pattern. Initialization is used only at the beginning of the algorithm. However that part is not subject to evolution.}
\item{$Selection$. In the MEP-based approach used in \cite{oltean_ecal} we have used a selection operator which has 2 fixed parameters (e.g. $Selection$ 5, 2). In the current approach we use a more general type of selection operator which has no parameter. This operator also selects two random individuals and outputs the best of them, but is not bound anymore to some fixed positions in the population.}
\item{$Crossover$ - no modifications.}
\item{$Mutation$ - no modifications.}

\end{itemize}

\subsection{Decoding MEP individuals}

The individual generated by the last instruction of the MEP chromosome will be added to the new population. It can be easily seen that not all MEP instructions are effective. Thus, we have to see the sequence of instructions that has generated that individual. In order to extract all effective instructions we proceed in a bottom-up manner. We begin by marking the last instruction in the chromosome and then we follow the function pointers recursively until we get some terminals. The marked MEP genes are the only useful genetic instructions for our purpose.

For instance in the example 1 only the instructions in positions 2, 4 and 5 are actually utilized. Thus the pattern will be reduced to 3 instructions only:\\

1: $Selection$

2: $Mutation$ 1

3: $Crossover$ 1, 2\\

The other instructions (introns) have been removed from this final pattern and the function pointers have been relabeled in order to reflect changes. Note that the introns removal is performed only for the best chromosome (in the population) at the end of the search process.\\

\textbf{Remark.}

Other operators could be added to the function and terminal sets in order to define a more complex behavior for the evolved pattern. For instance we could add an operator that computes the worst individual in the current population. This operator is very useful in a steady-state evolutionary model \cite{syswerda1}.

\subsection{Where the pattern is embedded?}\label{embedded_alg}

\begin{figure*}[htbp]
\centerline{\includegraphics[width=\textwidth]{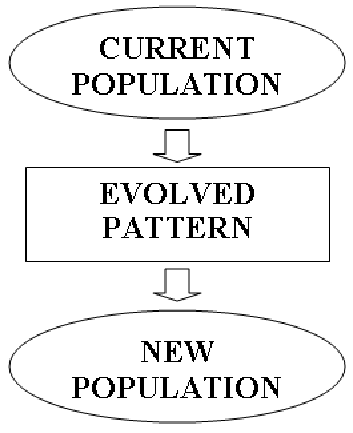}} 
\caption{The dataflow view of the process of obtaining new individuals using the evolved pattern. The input for the pattern is the current population. The output (the individual obtained by applying the pattern) is always added to the new population.}
\label{use_pattern}
\end{figure*}

The evolved pattern is a sequence of instructions which will repeatedly be used for obtaining new individuals. This process is depicted in Figure \ref{use_pattern}. Note that the proposed approach tells us only the way in which a new individual is generated based on the existing individuals (the current population) and it does not tell us what to do with the obtained individual. In this model we have chosen to add the obtained individual to the new population. The new population will replace the current population at the end of a generation and the search process continues. 

The algorithm where the pattern will be embedded is a standard one: it starts with a random population of solutions; always copies the best-of-generation individual in the next generation and all the other individuals are obtained by applying the pattern.

The evolutionary algorithm embedding the pattern is given below:

\begin{center}
\textbf{The Evolutionary Algorithm embedding the Evolved Pattern}
\end{center}

\textsf{S}$_{1}$\textsf{. Randomly create the initial population }\textsf{\textit{P}}\textsf{(0)}

\textsf{S}$_{2}$\textsf{. }\textsf{\textbf{for}}\textsf{ 
}\textsf{\textit{t}}\textsf{ = 1 }\textsf{\textbf{to}}\textsf{ 
}\textsf{\textit{Max}}\textsf{{\_}}\textsf{\textit{Generations}}\textsf{ 
}\textsf{\textbf{do}}

\textsf{S}$_{3}$\textsf{. 
}\hspace{0.5cm}\textsf{\textit{P'}}\textsf{(}\textsf{\textit{t}}\textsf{) = \{ The best individual in $P(t)$ \};}

\hspace{1.2cm}\textsf{//The best individual in the current population is added to the new population}

\textsf{S}$_{4}$\textsf{.}\hspace{0.5cm}\textsf{\textbf{for}}\textsf{ 
}\textsf{\textit{k}}\textsf{ = 2 }\textsf{\textbf{to}}\textsf{ $\vert 
$}\textsf{\textit{P}}\textsf{(}\textsf{\textit{t}}\textsf{)$\vert $ 
}\textsf{\textbf{do}}

\textsf{S}$_{5}$\textsf{.}\hspace{1cm}\textsf{RUN THE PATTERN()  // the evolved pattern is run here } 

\hspace{5.5cm}\textsf{// an offspring }\textsf{\textit{offspr}}\textsf{ is obtained}

\textsf{S}$_{6}$\textsf{.}\hspace{1cm}\textsf{Add }\textsf{\textit{offspf}}\textsf{ to 
}\textsf{\textit{P'}}\textsf{(}\textsf{\textit{t}}\textsf{); //add 
}\textsf{\textit{offspr} in the new population}

\textsf{S}$_{7}$\textsf{.}\hspace{0.5cm}\textsf{\textbf{endfor}}

\textsf{S}$_{8}$\textsf{.}\hspace{0.5cm}\textsf{\textit{P}}\textsf{(}\textsf{\textit{t}}\textsf{+1) = 
}\textsf{\textit{P}}\textsf{'(}\textsf{\textit{t}}\textsf{);}

\textsf{S}$_{9}$\textsf{. }\textsf{\textbf{endfor}}\\

Note that this algorithm will be used (see section \ref{use_the_evolved}) for solving particular problems such as function optimization. After obtaining the pattern we will not be anymore interested in MEP and other techniques for evolving complex computer programs (such EAs). We will focus only in using the pattern for solving some particular problems.

\subsection{Fitness Assignment}

We deal with EAs at two different levels: a micro level representing the 
pattern encoded into an MEP chromosome and a macro level GA, 
which evolves MEP individuals. Macro level GA execution is bounded by known 
rules for GAs (see \cite{goldberg1}). 

In order to compute the fitness of an MEP individual we have to compute the quality 
of the pattern encoded in that chromosome. For this purpose, the pattern encoded into a 
MEP chromosome is embedded into an evolutionary algorithm which is run on the particular problem being solved (as shown in the previous section). A solution (for the particular problem being solved) is obtained. For instance, this solution could be a real-valued array (in the case of function optimization problems) or could be a path in a graph in the case of the TSP problem.

Roughly speaking, the fitness of an MEP individual equals the fitness of 
the solution generated by the pattern encoded into that 
MEP chromosome. But, since the pattern encoded into an MEP chromosome uses 
pseudo-random numbers, it is very likely that successive runs of the same EA 
will generate completely different solutions. This stability problem is 
handled in a standard manner: the EA embedding the pattern encoded into an MEP chromosome is 
executed (run) more times (100 runs are, in fact, performed in all the 
experiments for evolving EAs for function optimization) and the fitness of an MEP chromosome is the 
average of the fitness of the EA encoded in that chromosome over all runs.

The optimization type (minimization/maximization) of the macro level EA is the same as the optimization type of the micro level EA. In our experiments we have employed a minimization relation (finding the minimum of a function).

\subsection{The Model used for Evolving EAs}

We use the steady state algorithm described in section \ref{algorithm} in order to evolve EAs. For increasing the generalization ability (i.e. the ability of the evolved EA to yield good results on new test problems), the problem set 
is divided into three sets, suggestively called training set, 
validation set and test set (see \cite{prechelt1}). In our experiments the training set 
consists in a difficult test problem. Validation is performed by using another 
difficult test problem. The test set consists in some other well-known 
benchmarking problems.

A method called \textit{early stopping} \cite{prechelt1} is used in order to avoid the overfitting of the population individuals to the particular training examples used. This method consists in 
computing the test set performance for the chromosome which had the minimum 
validation error during the search process. The use of early stopping 
technique will increase the generalization performance \cite{prechelt1}.

The test set consists in several well-known benchmarking problems \cite{burkard1,reinelt1,yao1} 
used for assessing the performances of the evolutionary algorithms.

\subsection{The Algorithm for Evolving EAs}\label{algorithm}

For evolving MEP individuals we use a steady-state \cite{syswerda1} evolutionary algorithm. The sketch of this algorithm has been depicted in Figure \ref{fig_mep_eas}.

The steady-state MEP algorithm starts with a randomly chosen population of 
individuals. The following steps are repeated until a termination condition 
is reached: Two parents are selected by using binary tournament 
and then they are recombined with a fixed crossover probability. Two offspring are obtained by the recombination 
of two parents. The offspring are mutated and the 
better of them replaces the worst individual in the current population (if the 
offspring is better than the worst individual in the current population).

The output of the MEP algorithm is the best evolutionary pattern encoded into a MEP chromosome. This pattern (sequence of intructions) will be embedded in the algorithm described in section \ref{embedded_alg}. The obtained algorithm will be used in section \ref{use_the_evolved} for solving several well-known optimization problems.

\subsection{Complexity and running time}

The complexity of the proposed method is bounded by the known rules for evolutionary algorithm. In the case of the algorithm where evolved pattern is embeded (see section \ref{embedded_alg}) then complexity can be simply described by the classic equation:

\begin{equation}
\label{complex_embeded}
CA = O(PopSize * NumberOfGenerations * NumberOfGenes * C),
\end{equation}

where $NumberOfGenes$ is the number of genes in the MEP pattern, $C$ is the complexity induced by the representation (real-valued array, TSP path etc) and the other parameters are self explaining.

The complexity of the algorithm used for evolving patterns is given by the formula:

\begin{equation}
\label{complex_macro}
O(PopSize * NumberOfGenerations * CA).
\end{equation}

The $CA$ factor was introduced here because we need to compute the fitness of each newly evolved pattern, which actually means that we have to run the algorithm given in section \ref{embedded_alg} whose complexity is described by equation \ref{complex_embeded}.

Note that the values of parameters $PopSize$ and $NumberOfGenerations$ may be different in formulas \ref{complex_embeded} and \ref{complex_macro}.

The process of evolving algorithms is a complex task which requires many computational resources. This is because we need to assess the performance of each evolved pattern by applying it to a particular problem (function optimization in our case). For evolving an evolutionary pattern (see section \ref{experiments}) we need about 1 hour. This is a significant improvement compared to the technique proposed in \cite{oltean_mit} which requires about 1 day to evolve an algorithm.

\section{Numerical Experiments}\label{experiments}

In this section, several numerical experiments for evolving EAs are 
performed. An evolutionary algorithm for function optimization is evolved. Several numerical experiments, with a standard Genetic Algorithm \cite{goldberg1} for function optimization, are also performed in order to assess the performance of the evolved EA. Finally the results are compared.

\subsection{Evolving EAs for Function Optimization}\label{funcopt}

In this section, an Evolutionary Algorithm for function optimization is evolved.

\subsubsection{Test Functions}

Ten test problems $f_{1}-f_{10}$ (given in Table \ref{tab_test_functions}) are used in order to asses the 
performance of the evolved EA. Functions $f_{1}-f_{6}$ are unimodal test 
function. Functions $f_{7}-f_{10}$ are highly multimodal (the number of the
local minima increases exponentially with the problem dimension \cite{yao1}).

\begin{table*}[htbp]
\begin{center}
\caption{Test functions used in our experimental study. The parameter $n$ is the space dimension ($n$ = 5 in our numerical experiments) and $f_{min}$ is the minimum value of the function.}
\label{tab_test_functions}
\begin{tabular}
{|p{220pt}|p{80pt}|p{55pt}|}
\hline
Test function& 
Domain& 
$f_{min}$ \\
\hline
$f_1 (x) = \sum\limits_{i = 1}^n {(i \cdot x_i^2 )} .$& 
[-10, 10]$^{ n}$& 
0 \\
\hline
$f_2 (x) = \sum\limits_{i = 1}^n {x_i^2 } .$& 
[-100, 100]$^{ n}$& 
0 \\
\hline
$f_3 (x) = \sum\limits_{i = 1}^n {\vert x_i \vert + \prod\limits_{i = 1}^n {\vert x_i \vert } } .$& 
[-10, 10]$^{ n}$& 
0 \\
\hline
$f_4 (x) = \sum\limits_{i = 1}^n {\left( {\sum\limits_{j = 1}^i {x_j} } \right)^2} .$& 
[-100, 100$^{ }$]$^{ n}$& 
0 \\
\hline
$f_5 (x) = \max _i \{x_i ,1 \le i \le n\}.$& 
[-100, 100]$^{ n}$& 
0 \\
\hline
$f_{6} (x) = \sum\limits_{i = 1}^{n - 1} {100 \cdot (x_{i + 1} - x_i^2 )^2 + (1 - x_i )^2} .$& 
[-30, 30]$^{ n}$& 
0 \\
\hline
$f_7 (x) = 10 \cdot n + \sum\limits_{i = 1}^n {(x_i^2 - 10 \cdot \cos (2 \cdot \pi \cdot x_i ))} $& 
[-5, 5]$^{ n}$& 
0 \\
\hline
$f_8 (x) = - a \cdot e^{ - b\sqrt {\frac{\sum\limits_{i = 1}^n {x_i^2 } }{n}} } - e^{\frac{\sum {\cos (c \cdot x_i )} }{n}} + a + e.$& 
[-32, 32]$^{ n}$ \par $a$ = 20, $b$ = 0.2, $c$ = 2\textit{$\pi $}.& 
0 \\
\hline
$f_9 (x) = \frac{1}{4000} \cdot \sum\limits_{i = 1}^n {x_i^2 - \prod\limits_{i = 1}^n {\cos (\frac{x_i }{\sqrt i }) + 1} } .$& 
[-500, 500]$^{ n}$& 
0 \\
\hline
$f_{10} (x) = \sum\limits_{i = 1}^n {( - x_i \cdot \sin (\sqrt {\left| {x_i } \right|} ))} $& 
[-500, 500]$^{ n}$& 
-$n * $ 418.98 \\
\hline
\end{tabular}
\end{center}
\end{table*}

\subsubsection{Experimental Results}\label{ga}

In this section we evolve an EA for function optimization and then we asses 
its performance. A comparison with standard GA is 
performed further in this section.

For evolving an EA we use $f_{1}$ as the training problem and 
$f_{2}$ as the validation problem. The number of dimensions was set to 5 for all numerical experiments.

An important issue concerns the solutions evolved by the EAs encoded into an 
MEP chromosome and the specific genetic operators used for this purpose. 
The solutions evolved by the EA encoded into MEP chromosomes are represented 
by using real values \cite{goldberg1}. Thus, each chromosome of the evolved EA is a fixed-length array of real values. By initialization, a point within the definition 
domain is randomly generated. Convex crossover with $\alpha $ = 
$\raise.5ex\hbox{$\scriptstyle 1$}\kern-.1em/ 
\kern-.15em\lower.25ex\hbox{$\scriptstyle 2$} $ and Gaussian mutation with 
$\sigma $ = 0.01 are used \cite{goldberg1}.

A short description of real encoding and the corresponding genetic operators is given in Table \ref{real}.

\begin{table*}[htbp]
\begin{center}
\caption{A short description of real encoding.}
\label{real}
\begin{tabular}
{|p{150pt}|p{200pt}|}
\hline
Function to be optimized& 
$f$:[\textit{MinX}, \textit{MaxX}]$^{n} \to \Re $ \\
\hline
Individual representation& 
$x$ = ($x_{1}$, $x_{2}$, \ldots , $x_{n})$. \\
\hline
Convex Recombination with $\alpha $ = 0.5& 
parent 1 -- $x$ = ($x_{1}$, $x_{2}$, \ldots , $x_{n})$. \par parent 2 -- $y$ = ($y_{1}$, $y_{2}$, \ldots , $y_{n})$. \par the offspring -- $o = (\frac{x_1 + y_1 }{2},\,\frac{x_2 + y_2 }{2},\,...\,,\,\frac{x_n + y_n }{2}).$ \\
\hline
Gaussian Mutation& 
the parent -- $x$ = ($x_{1}$, $x_{2}$, \ldots , $x_{n})$. \par the offspring -- $o$ = ($x_{1 }+G$(0,$\sigma )$, $x_{2 }+G$(0,$\sigma )$, \ldots , $x_{n }+G$(0,$\sigma ))$, \par where $G$ is a function that generates real values with Gaussian distribution. \\
\hline
\end{tabular}
\end{center}
\end{table*}

\textbf{Experiment 1}\\

An Evolutionary Algorithm for function optimization is evolved in this experiment.

The parameters of the MEP algorithm are given in Table \ref{tab3}.

\begin{table*}[htbp]
\begin{center}
\caption{The parameters of the MEP algorithm used for evolving evolutionary algorithms.}
\label{tab3}
\begin{tabular}
{|p{140pt}|p{171pt}|}
\hline
\textbf{Parameter}& 
\textbf{Value} \\
\hline
Population size& 
100 \\
\hline
Code Length& 
15 instructions \\
\hline
Number of generations& 
100 \\
\hline
Crossover probability& 
0.8 \\
\hline
Crossover type& 
Uniform Crossover \\
\hline
Mutation & 
1 mutation per chromosome \\
\hline
Function set& 
$F$ = {\{}\textit{Crossover}, \textit{Mutation}{\}} \\
\hline
Terminal set& 
$F$ = {\{}\textit{Selection}{\}} \\
\hline
\end{tabular}
\end{center}
\end{table*}

The parameters of the evolved EA are given in Table \ref{tab4}.

\begin{table*}[htbp]
\begin{center}
\caption{The parameters of the evolved EA for function optimization.}
\label{tab4}
\begin{tabular}
{|p{140pt}|p{170pt}|}
\hline
\textbf{Parameter}& 
\textbf{Value} \\
\hline
Individual representation&
fixed-length array of real values.\\
\hline
Population size& 
50 \\
\hline
Number of generations& 
50 \\
\hline
Crossover probability& 
1 \\
\hline
Crossover type& 
Convex Crossover with $\alpha $ = 0.5 \\
\hline
Mutation & 
Gaussian mutation with $\sigma $ = 0.01  \\
\hline
Mutation probability&
1\\
\hline
Selection& 
Binary Tournament \\
\hline
\end{tabular}
\end{center}
\end{table*}

\begin{figure*}[htbp]
\centerline{\includegraphics[width=\textwidth]{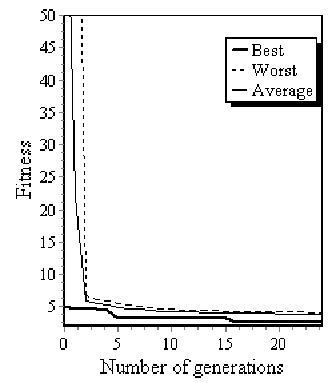}} \caption{The evolution of the fitness of the best individual, the worst individual and the average fitness in a particular run. We have depicted a window of 25 runs just in order to provide a better visualisation.}
\label{fig1}
\end{figure*}

The results of this experiment are depicted in Figure \ref{fig1}.

The effectiveness of our approach can be seen in Figure \ref{fig1}. The MEP 
technique is able to evolve an EA for solving optimization problems. The 
quality of the evolved EA improves as the search process advances. 

One complete run (100 generations) took about 1 hour. Note that one run in the case of the LGP approach \cite{oltean_lgp} took one day.

One of the best evolved evolutionary patterns is given below in MEP notation. Note that we have removed the introns \cite{brameier1} and we have kept only the effective code. This is why the pattern is shorter than the MEP chromosome used in evolution (see Table \ref{tab3}).\\

1: $Selection$

2: $Selection$

3: $Mutation$ 2

4: $Mutation$ 3

5: $Crossover$ 1, 4

6: $Mutation$ 5

7: $Mutation$ 6

8: $Mutation$ 7

9: $Mutation$ 8\\

This chromosome selects two individuals ($a$ and $b$), mutates twice the individual $b$ and then performs a crossover between the result and $a$. Then the offspring is mutated four times. The structure of the chromosome suggests that the standard mutation (with $\sigma$ = 0.01) is not sufficient for this test function. Additional mutations are performed in order to improve the algorithm's performance.\\

We have also analyzed the quality of other individuals involved in the search process. We have randomly chosen one of the runs and we have analyzed the fitness of the best and of the worst individual in the population (see Figure \ref{fig1}). The fitness of the best individual in the first generation is 4.92, while the average fitness (of all the individuals at generation 0) is 85.15. This means that we have some good individuals in the first generation, but we also have some very unfit individuals. The worst individual in the first generation has the fitness equal to 151.70. This individual is given below in MEP notation:\\

1: $Selection$

2: $Mutation$ 1

3: $Selection$

4: $Crossover$ 2, 3

5: $Selection$

6: $Crossover$ 4, 5

7: $Selection$

8: $Crossover$ 6, 7

9: $Mutation$ 8 \\

In the second generation the average fitness is 22.01 and the fitness of the worst individual is 121.02. In the third generation the worst individual in the population has the fitness 6.83, whereas the fitness of the best individual is 4.06. We see that in very few generations the average fitness quickly decreases, getting close to the value of the best fitness in the population. This means that the very unfit individuals are quickly replaced by some very fit individuals.\\

\textbf{Experiment 2}\label{use_the_evolved}\\

This experiment serves our purpose of comparing the evolved EA with a standard Genetic 
Algorithm. The parameters used by the evolutionary algorithm embedding the evolved pattern are given in Table \ref{tab4}. The parameters used by standard GA are given in 
Table \ref{tab2}. The results of the comparison are given in Table \ref{tab_ga}.

The standard GA algorithm, used for comparison, is given below:

\begin{center}
\textbf{Standard GA algorithm}
\end{center}

\textsf{S}$_{1}$\textsf{. Randomly create the initial population }\textsf{\textit{P}}\textsf{(0)}

\textsf{S}$_{2}$\textsf{. }\textsf{\textbf{for}}\textsf{ 
}\textsf{\textit{t}}\textsf{ = 1 }\textsf{\textbf{to}}\textsf{ 
}\textsf{\textit{Max}}\textsf{{\_}}\textsf{\textit{Generations}}\textsf{ 
}\textsf{\textbf{do}}

\textsf{S}$_{3}$\textsf{. 
}\hspace{0.5cm}\textsf{\textit{P'}}\textsf{(}\textsf{\textit{t}}\textsf{) = \{ The best individual in $P(t)$ \};}

\textsf{S}$_{4}$\textsf{.}\hspace{0.5cm}\textsf{\textbf{for}}\textsf{ 
}\textsf{\textit{k}}\textsf{ = 2 }\textsf{\textbf{to}}\textsf{ $\vert 
$}\textsf{\textit{P}}\textsf{(}\textsf{\textit{t}}\textsf{)$\vert $ 
}\textsf{\textbf{do}}

\textsf{S}$_{5}$\textsf{.}\hspace{1cm}\textsf{\textit{p}}$_{1}$\textsf{ = 
}\textsf{\textit{Selection}}\textsf{(}\textsf{\textit{P}}\textsf{(}\textsf{\textit{t}}\textsf{)); 
// select an individual from the population}

\textsf{S}$_{6}$\textsf{.}\hspace{1cm}\textsf{\textit{p}}$_{2}$\textsf{ = 
}\textsf{\textit{Selection}}\textsf{(}\textsf{\textit{P}}\textsf{(}\textsf{\textit{t}}\textsf{)); 
// select the second individual }

\textsf{S}$_{7}$\textsf{.}\hspace{1cm}\textsf{\textit{Crossover}}\textsf{ 
(}\textsf{\textit{p}}$_{1}$\textsf{, }\textsf{\textit{p}}$_{2}$\textsf{, 
}\textsf{\textit{offsp}}\textsf{); // crossover the parents p}$_{1}$\textsf{ 
and p}$_{2}$

\hspace{1.5cm}\textsf{// an offspring }\textsf{\textit{offspr}}\textsf{ is obtained}

\textsf{S}$_{8}$\textsf{.}\hspace{1cm}\textsf{\textit{Mutation}}\textsf{ 
(}\textsf{\textit{offspr}}\textsf{); // mutate the offspring 
}\textsf{\textit{offspr}}

\textsf{S}$_{9}$\textsf{.}\hspace{1cm}\textsf{ Add }\textsf{\textit{offspf}}\textsf{ to 
}\textsf{\textit{P'}}\textsf{(}\textsf{\textit{t}}\textsf{); //move 
}\textsf{\textit{offspr} in the new population}

\textsf{S}$_{10}$\textsf{.}\hspace{0.5cm}\textsf{\textbf{endfor}}

\textsf{S}$_{11}$\textsf{.}\hspace{0.5cm}\textsf{\textit{P}}\textsf{(}\textsf{\textit{t}}\textsf{+1) = 
}\textsf{\textit{P}}\textsf{'(}\textsf{\textit{t}}\textsf{);}

\textsf{S}$_{12}$\textsf{. }\textsf{\textbf{endfor}}\\

The best solution in the last generation is the output of the program.

The pattern (the sequence of code that is repeatedly used) is given by steps $S_5-S_9$.

The parameters of the standard GA are given in Table \ref{tab2}. Results are given in Table \ref{tab_ga}.

\begin{table*}[htbp]
\begin{center}
\caption{The parameters of a standard GA for Experiment 2.}
\label{tab2}
\begin{tabular}
{|p{120pt}|p{190pt}|}
\hline
\textbf{Parameter}& 
\textbf{Value} \\
\hline
Population size& 
50\\
\hline
Individual encoding& 
fixed-length array of real values\\
\hline
Number of generations& 
50 \\
\hline
Crossover probability& 
1 \\
\hline
Crossover type& 
Convex Crossover with $\alpha $ = 0.5 \\
\hline
Mutation & 
Gaussian mutation with $\sigma $ = 0.01  \\
\hline
Mutation probability&
1\\
\hline
Selection& 
Binary Tournament \\
\hline
\end{tabular}
\end{center}
\end{table*}

\begin{table*}[htbp]
\begin{center}
\caption{The results obtained by applying the Evolved EA and the Standard GA to the considered test functions. StdDev stands for the standard deviation. The results are averaged over 500 runs.}
\label{tab_ga}
\begin{tabular}
{|p{60pt}|p{40pt}|p{40pt}|p{40pt}|p{40pt}|p{40pt}|p{40pt}|p{40pt}|p{40pt}|}
\hline
\raisebox{-1.50ex}[0cm][0cm]{\textbf{Test function}}& 
\multicolumn{4}{|p{160pt}|}{\textbf{Evolved EA}} & 
\multicolumn{4}{|p{160pt}|}{\textbf{Standard GA}}  \\
\cline{2-9}
 & 
Best run&
Worst run&
Mean& 
StdDev& 
Best run&
Worst run&
Mean& 
StdDev \\
\hline
$f_{1}$& 
2.62E-5&
4.306&
0.128&
0.401&
1.424E-5&
5.503&
0.335&
0.691\\
\hline
$f_{2}$& 
6.878E-5&
366.563&
27.115&
38.991&
3.990E-5&
468.196&
25.963&
35.553 \\
\hline
$f_{3}$& 
4.852E-3&
2.297&
0.250&
0.364&
3.676E-3&
4.708&
0.487& 
0.571 \\
\hline
$f_{4}$& 
6.244E-5&
458.838&
38.277&
58.218&
4.899E-4&
417.656&
38.821& 
57.073 \\
\hline
$f_{5}$& 
6.592E-3&
14.279&
2.996&
2.306&
3.283E-3&
14.701&
3.032&
2.120 \\
\hline
$f_{6}$& 
5.513E-2&
9801.02&
339.818&
993.006&
4.310E-3&
27823.4&
485.559&
1935.170 \\
\hline
$f_{7}$& 
3.073E-3&
15.361&
1.861&
1.577&
4.636E-3&
19.304&
3.243&
3.811 \\
\hline
$f_{8}$& 
1.036E-2&
8.536&
2.775&
1.756&
4.915E-3&
9.507&
2.905&
1.599 \\
\hline
$f_{9}$& 
2.163E-2&
3.520&
0.509&
0.340&
2.123E-2&
3.387&
0.533&
0.323 \\
\hline
$f_{10}$& 
-1641.83&
-569.444&
-1010.610&
171.799&
-1642.01&
-610.59&
-998.549&
174.782 \\
\hline
\end{tabular}
\end{center}
\end{table*}

Taking into account the averaged values we can see in table \ref{tab_ga} that the evolved EA significantly performs better than the standard GA in 9 cases (out of 10). For the test function $f_{2}$ the latter performs better than the evolved pattern. 

When taking into account the solution obtained in the best run, the standard GA performs better than the evolved EA in 8 cases (out of 10).

When taking into account the solution obtained in the worst run, the evolved EA performs better than the standard GA in 7 cases (out of 10).

In order to determine whether the differences between the Evolved EA and the standard GA are statistically 
significant, we use a $t$-test with 95{\%} confidence. Only the best solutions in each run have been taken into account for these tests. Before applying the $t$-test, an $F$-test has been used for determining whether the compared data have the 
same variance. The $P$-values of a two-tailed $t$-test are given in Table \ref{tab_ttest}.

\begin{table*}[htbp]
\begin{center}
\caption{The results of the t-Test and F-Test with 499 degrees of freedom.}
\label{tab_ttest}
\begin{tabular}
{|p{50pt}|p{54pt}|p{52pt}|}
\hline
Function& 
F-Test& 
t-Test \\
\hline
$f_{1}$& 
9.22E-10& 
8.59E-10 \\
\hline
$f_{2}$& 
3.94E-2& 
6.25E-1 \\
\hline
$f_{3}$& 
5.08E-23& 
1.72E-14 \\
\hline
$f_{4}$& 
6.57E-1& 
8.81E-1 \\
\hline
$f_{5}$& 
5.97E-2& 
7.96E-1 \\
\hline
$f_{6}$& 
5.65E-47& 
1.34E-1 \\
\hline
$f_{7}$& 
2.75E-77& 
2.10E-13 \\
\hline
$f_{8}$&
3.69E-2&
2.21E-1 \\
\hline
$f_{9}$&
2.33E-1&
2.48E-1 \\
\hline
$f_{10}$& 
7.00E-1& 
2.71E-1 \\
\hline
\end{tabular}
\end{center}
\end{table*}

Table \ref{tab_ttest} shows that the difference between the EA embedding the evolved pattern and the 
standard GA is statistically significant ($P < 0.05$) for 6 test problems.\\

\textbf{Experiment 3}\\

We are also interested in analyzing the relationship between the number of 
generations of the evolved EA and the quality of the solutions obtained by 
applying the evolved EA to the considered test functions. The parameters of 
the Evolved EA are given in Table \ref{tab4} and the parameters of the standard 
GA are given in Table \ref{tab2}.

\begin{figure*}[htbp]
\centerline{\includegraphics[width=\textwidth]{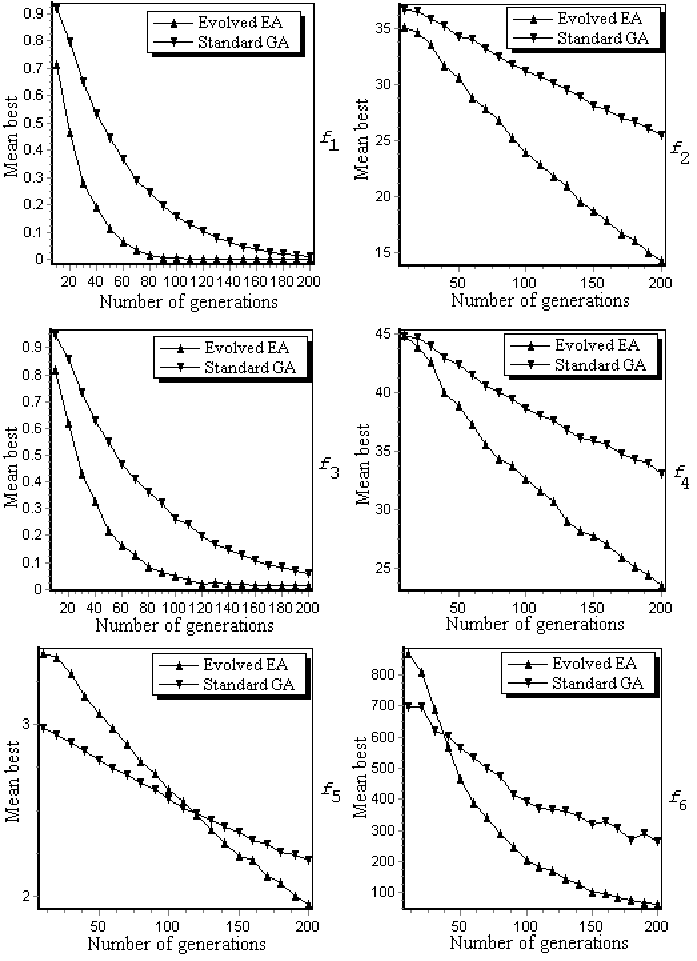}} 
\caption{The relationship between the number of generations and the quality of the solutions obtained by the Evolved EA and by the standard GA for the unimodal test functions $f_{1}-f_{6}$. The number of generations varies between 10 and 200. Results are averaged over 100 runs.}
\label{fig2}
\end{figure*}

\begin{figure*}[htbp]
\centerline{\includegraphics[width=\textwidth]{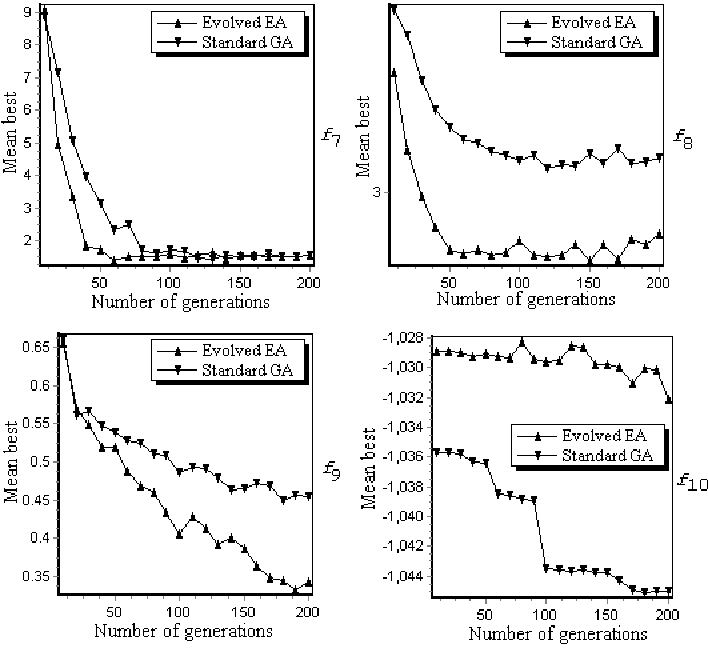}}
\caption{The relationship between the number of generations and the quality of the solutions obtained by the Evolved EA and by the standard GA for the multimodal test functions $f_{7}-f_{10}$. The number of generations varies between 10 and 200. Results are averaged over 100 runs.}
\label{fig3}
\end{figure*}

The results of this experiment are depicted in Figure \ref{fig2} (the unimodal 
test functions) and in Figure \ref{fig3} (the multimodal test functions).

Figures \ref{fig2} and \ref{fig3} show that the Evolved EA is scalable regarding the number of generations. We can see a continuous improvement tendency during the search process for all test functions ($f_{1}-f_{10})$.\\

\textbf{Experiment 4}\\

We are also interested in analyzing the relationship between the population size of the evolved EA and the quality of the solutions obtained by 
applying the evolved EA to the considered test functions. The parameters of 
the Evolved EA are given in Table \ref{tab4} and the parameters of the standard 
GA are given in Table \ref{tab2}.

\begin{figure*}[htbp]
\centerline{\includegraphics[width=\textwidth]{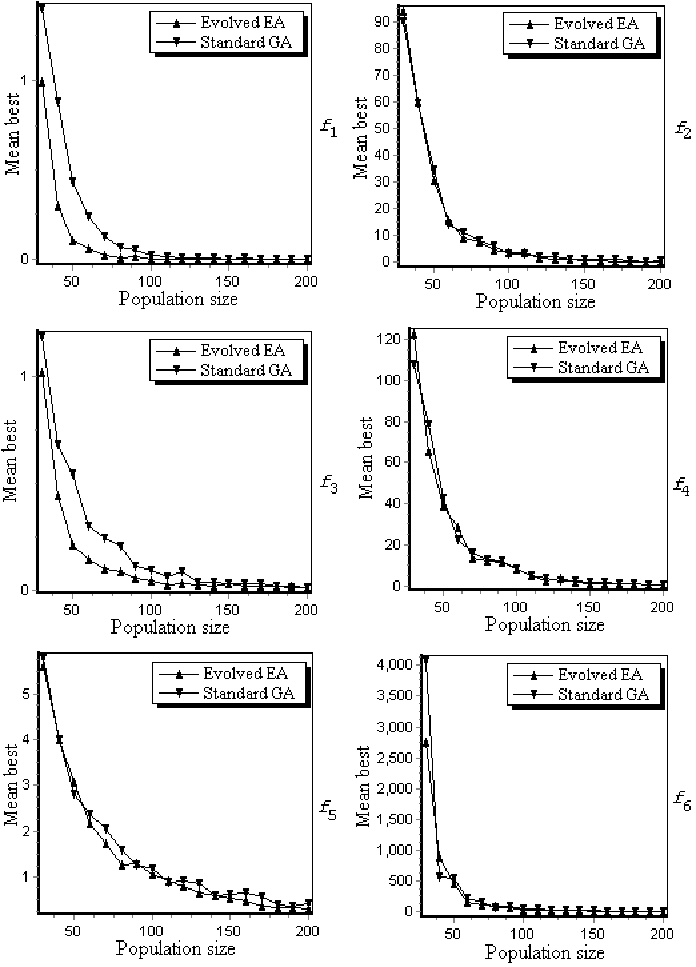}} 
\caption{The relationship between the population size and the quality of the solutions obtained by the Evolved EA and by the standard GA for the unimodal test functions $f_{1}-f_{6}$. Population size varies between 30 and 200. Results are taken from the last generation and averaged over 100 independent runs.}
\label{fig4}
\end{figure*}

\begin{figure*}[htbp]
\centerline{\includegraphics[width=\textwidth]{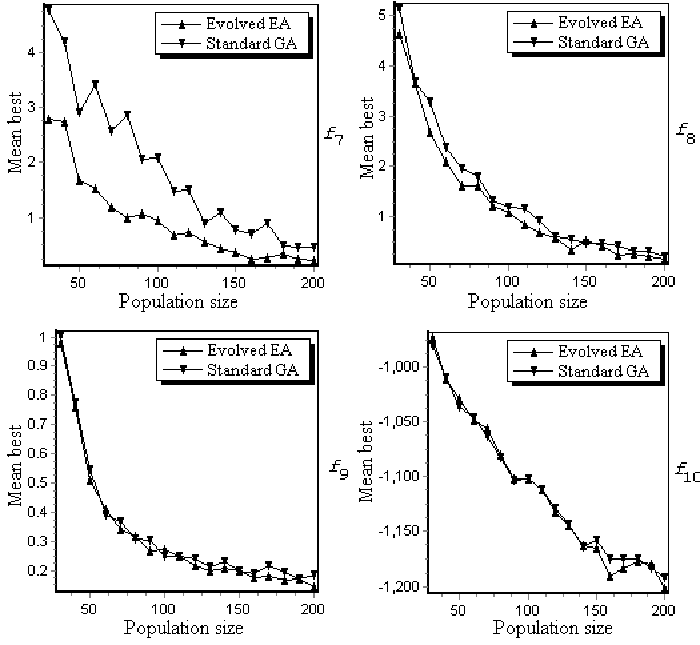}} \caption{The relationship between the population size and the quality of the solutions obtained by the Evolved EA and by the standard GA for the multimodal test functions $f_{7}-f_{10}$. Population size varies between 30 and 200. Results are averaged over 100 runs.}
\label{fig5}
\end{figure*}

The results of this experiment are depicted in Figure \ref{fig4} (the unimodal 
test functions) and in Figure \ref{fig5} (the multimodal test functions).

Figures \ref{fig4} and \ref{fig5} show that the Evolved EA is scalable on what concerns the number of individuals in the population. We can see a continuous improvement tendency during the search process for all test functions ($f_{1}-f_{10})$.

\section{Conclusions and further work}\label{future}

A new model for evolving Evolutionary Algorithms has been proposed in this paper. A detailed description of the proposed approach has been given, thus allowing researchers to apply the method for evolving 
Evolutionary Algorithms that could be used for solving problems in their 
fields of interest. 

The proposed model has been used for evolving Evolutionary Algorithms for function optimization. Numerical experiments emphasize the robustness and the efficacy of this approach. The evolved Evolutionary Algorithms perform similarly and sometimes even better than some standard approaches in the literature for the considered test functions.

In order to evolve high quality EAs and assess their performance, an 
extended set of training problems should be used. This set should include 
problems from different fields, such as function optimization, symbolic 
regression \cite{koza1}, the Traveling Salesman Problem \cite{freisleben1,garey1}, the Quadratic Assignment Problem \cite{krasnogor1,merz2} classification etc. Further efforts will be dedicated to 
the training of such an algorithm which should have an increased generalization 
ability.

An extended set of operators will be used in order to obtain more powerful Evolutionary Algorithms. This set will include operators that compute the fitness of the best/worst individual in the population. In this case, the evolved EA will have the "elitist" feature which will allow us to compare it with more complex evolutionary schemes like steady-state \cite{syswerda1}.

For obtaining better results we also plan to evolve the parameters involved by the pattern (e.g. mutation probability, crossover probability etc). In this case, the evolved structure will contain a MEP chromosome and some numerical values for the parameters whose optimal value we want to find.

Only fixed-size populations have been used in our experiments. Another 
extension of the proposed approach will take into account the scalability of 
the population size.

Further effort will be focused on evolving evolutionary algorithms for other difficult (NP-Complete problem) \cite{burkard1,garey1,reinelt1}.

\end{document}